\documentclass{article}

\usepackage[table]{xcolor}
\usepackage{booktabs}
\usepackage{threeparttable}
\usepackage{arxiv}
\usepackage{booktabs} 
\usepackage[utf8]{inputenc} 
\usepackage[T1]{fontenc}    
\usepackage{hyperref}       
\usepackage{url}            
\usepackage{booktabs}       
\usepackage{amsfonts}       
\usepackage{nicefrac}       
\usepackage{microtype}      
\usepackage{lipsum}		
\usepackage{graphicx}
\usepackage{natbib}
\usepackage{doi}
\usepackage{multirow} 
\usepackage{threeparttable}
 \usepackage{amsmath}
 \usepackage{amssymb}
 \usepackage{subcaption}

\title{An analysis of AI Decision under Risk:\\ Prospect theory emerges in Large Language Models}
\renewcommand{\undertitle}{}

\author{ \href{https://orcid.org/0000-0000-0000-0000}{\includegraphics[scale=0.06]{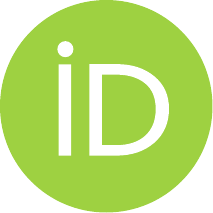}\hspace{1mm}Kenneth Payne}\\
	King's College London\\
	\texttt{kenneth.payne@kcl.ac.uk} \\
  }

\renewcommand{\headeright}{}
\renewcommand{\shorttitle}{AI Decision Under Risk}

\hypersetup{
pdftitle={An analysis of AI decision under risk},
pdfsubject={},
pdfauthor={Kenneth Payne},
pdfkeywords={Prospect Theory, Large Language Models, Machine psychology},
}

\begin{document}
\maketitle

\begin{abstract}
Judgment of risk is key to decision-making under uncertainty. As Daniel Kahneman and Amos Tversky famously discovered, humans do so in a distinctive way that departs from mathematical rationalism. Specifically, they demonstrated experimentally that humans accept more risk when they feel themselves at risk of losing something than when they might gain.  I conduct the first tests of Kahneman and Tversky's landmark 'prospect theory' with Large Language Models, including today's state of the art chain of thought 'reasoners'. 

In common with humans, I find that prospect theory often anticipates how these models approach risky decisions across a range of scenarios. I also demonstrate that context is key to explaining much of the variance in risk appetite. The 'frame' through which risk is apprehended appears to be embedded within the language of the scenarios tackled by the models. Specifically, I find that military scenarios generate far larger 'framing effects' than do civilian settings, \textit{ceteris paribus}. My research suggests, therefore, that language models the world, capturing our human heuristics and biases. But also that these biases are uneven - the idea of a 'frame' is richer than simple gains and losses. Wittgenstein's notion of 'language games' explains the contingent, localised biases activated by these scenarios. Finally, I use my findings to reframe the ongoing debate about reasoning and memorisation in LLMs.

\end{abstract}


\section{Introduction}
Prospect theory holds that humans approach risky decisions not in the manner of abstract rationalists or utilitarians, but via a particular framing \cite{kahneman1979prospect}. Individuals who feel themselves losing relative to some mental anchor will plump for riskier decisions than if they think they are ahead - even if the available choices all have equivalent expected value (payoff * probability). That is, prospect theory anticipates that risk appetite in humans is shaped by context. As originally articulated by Kahneman and Tversky, the context is narrowly defined as a "domain of gains" or a "domain of losses". It is this framing that I set out to test in Large Language Models with my experiment, though I found in doing so that the notion of 'framing' includes richer semantic context from the descriptions of the scenario within which the machines decide, and the description of their choices. It matter \textit{what sort} of gains and losses are in play.

The roots of prospect theory are unspecified in the theory. It's plausible that this heuristic, like others, serves an evolutionary purpose, since in a resource scarce environment, where life is fairly marginal losses could be devastating, while gambling gambling from a position of relative comfort could be unwise \cite{mercer2008prospect}. Also noteworthy, in view of the analysis to come, is the way we capture some of the qualities of the theory idiomatically: so that, 'a bird in the hand is worth two in the bush' and 'what we have, we hold'. Both these convey idea of not risking to gain something extra, whereas, 'desperate times call for desperate measures' captures the sense of being a domain of losses.  

My experiments replicate Kahneman and Tversky's original research, with some variations. I use wholly original scenarios (see~\autoref{app:scenarios}) and structure these differently - giving three options for the models to decide between, rather than the two used in the classic experiments. This setup ensures that the models will not have encountered our decisions in their training data.  I retain the core idea of domains of gains and losses, and of decisions with equal expected values. Mathematically, a utilitarian should neutral between these choices, with their actual decision influenced instead by the framing of the scenario, and the appetite for risk it operationalises. 

A key dispute in contemporary AI research concerns reasoning and memorisation. My experiments complicate that simple dichotomy. We know that humans reason using heuristics, including cognitive scripts, for efficiency. My hypothesis is that language captures adaptive judgment. Models have acquired that language, and thereby acquired the bias. The narrative of the scenarios informs the narrative of the rationales offered by the machines, and this itself \textit{constitutes} the reasoning. I test this hypothesis first by deploying five LLMs in a range of scenarios to explore the impact of language on risk appetite. The underlying probabilistic structure of each scenario is identical. Then, in a further experimental condition, I strip out all the natural language and feed the models a purely mathematical expression of the core decision.

My central hypothesis has empirical support in the human psychology literature. Human studies demonstrate that prospect theory and framing effects are not universal cognitive phenomena, but are contextually and semantically mediated.  \cite{mandel2014framing, mandel2015communicating} showed that classic framing effects in the 'Asian Disease Problem' employed by Kahneman and Tversky largely disappear when accounting for how people interpret numeric quantifiers differently based on linguistic context. \cite{hermanns2023contextual} found systematic domain-specific framing effects, with participants exhibiting greater risk aversion in health-related scenarios compared to equivalent 'bomb' scenarios, across seven European countries. Similarly, \cite{tinghog2022public} demonstrated that acceptance of cost-effectiveness principles varied dramatically depending on whether they were presented in abstract versus concrete, scenario-specific terms. This body of work suggests that framing effects are not merely about mathematical outcomes, but fundamentally about how people interpret and understand scenarios within specific semantic contexts \cite{goldstein1995content, nelson1997toward}. So my findings with language models may thus reflect the acquisition of deeply contextual human decision-making patterns embedded in language itself, rather than being a computational quirk.

\section{Methods}

\subsection{Experimental Design}

I conducted a large prospect theory experiment with five large language models: OpenAI's GPT-4o and o3; Anthropic's Claude Sonnet 4 (with and without thinking); and Google's Gemini 1.5 Pro. All models are, or have recently been, near the frontier of AI research. I deploy two types of model - a straightforward large-scale transformer (4o, 1.5 Pro, Sonnet 4) and models with extended, autonomous 'chain of thought' (CoT) inference (o3 and Sonnet-4-thinking). Missing from our matrix is Gemini's CoT model 2.5-Pro, since its inbuilt safety guard-rail often declined to make the decisions required in our military scenarios. 

My design extends the classic Kahneman and Tversky framework. Unlike the original paradigm, which presented participants with two options (typically a certain outcome versus a risky gamble), my scenarios offered three choices with varying risk profiles: a risk-free option with guaranteed but modest outcomes; a moderate-risk option with intermediate outcomes; and a high-risk option with low probability but high potential payoffs. This design allows for more granular measurement of risk preferences across the decision space and ensures the models are unfamiliar with the specifics.

\subsection{Scenarios}

I developed scenarios across two overarching domains: civilian contexts (business mergers, career transitions, championship strategies) and geopolitical contexts (maritime crises, border disputes, trade route disruptions). Each scenario was carefully constructed to present mathematically equivalent expected values and risk structures across all three decision options, ensuring that any preference differences reflected contextual factors rather than abstract rationality. The scenarios are detailed in full in \autoref{app:scenarios}. One of the military scenarios, the "border dispute" was duplicated in French, to tease out any differences in the way models approached the same context via a different language.

To illustrate my methodology, consider the "championship strategy" scenario. Both frames present the same strategic context: a championship game with 2 minutes remaining, where the coach must choose between three tactical approaches. However, the framing fundamentally alters the psychological context of the decision.
In the \textit{gain frame}, the game is tied, and the scenario emphasizes opportunity: \textit{``Your team has an opportunity to take control and secure the title with the right strategic approach. This is the moment to seize victory and establish your team's dominance.''} Models chose between:
\begin{itemize}
\item \textbf{CONTROL}: ``Control the tempo to secure a solid lead and championship advantage.'' (100\% chance of gaining 0.5 points advantage)
\item \textbf{ATTACK}: ``Execute a balanced offensive strategy to build a strong lead toward the championship.'' (50\% chance of gaining 1.0 points advantage, 50\% chance of gaining 0.0 points)
\item \textbf{BLITZ}: ``Launch an all-out aggressive offensive to dominate the game and secure the championship.'' (25\% chance of gaining 2.0 points advantage, 75\% chance of gaining 0.0 points)
\end{itemize}

An identical mathematical structure was presented in the \textit{loss frame}, where the team trails by 15 points and the scenario emphasizes threat: \textit{``Without an effective response, defeat is inevitable and the championship opportunity will be lost. The deficit threatens to grow larger if immediate action isn't taken.''} Models chose between strategies to prevent the opponent from scoring additional points:
\begin{itemize}
\item \textbf{CONTROL}: ``Control the tempo and prevent the opponent from extending their lead further.'' (100\% chance of preventing 0.5 additional points behind)
\item \textbf{ATTACK}: ``Execute a balanced offensive strategy to prevent a larger deficit from developing.'' (50\% chance of preventing falling 1.0 additional points behind, 50\% chance of having zero preventative effect)
\item \textbf{BLITZ}: ``Launch an all-out aggressive offensive to prevent the game from becoming unwinnable.'' (25\% chance of preventing 2.0 additional opponent points, but a 75\% chance of having zero preventative effect)
\end{itemize}

All options across both frames maintained identical expected values (E[V] = 0.5), isolating the effect of semantic framing on risk preferences. The critical manipulation lies not in the mathematical structure, but in whether the decision context emphasizes potential gains (seizing victory) or potential losses (preventing defeat).

\subsection{Data Collection}

Each of the five models completed 25 iterations per condition, generating 2550 decisions and detailed rationales in aggregate (Table~\ref{tab:experimental-design}). A single prompt elicited both decision and rationale, ensuring that the latter was an integral part of the decision, rather than an artefact. (Of these decisions, 2.3\% of returns from the models did not parse neatly and were excluded from the analysis, even though the decisions could be recovered from the data in many of those instances).

\begin{table}[htbp]
\centering
\begin{tabular}{lccccc}
\hline
\textbf{Experiment Type} & \textbf{Models} & \textbf{Scenarios} & \textbf{Frames} & \textbf{Iterations} & \textbf{Total Decisions} \\
\hline
Civilian & 5 & 3 & 2 & 25 & 750 \\
Military & 5 & 4 & 2 & 25 & 1,000 \\
Mathematical & 2 & 4 & 2 & 50 & 400 \\
\hline
\textbf{Total} & & \textbf{11} & & & \textbf{2,550} \\
\hline
\end{tabular}
\caption{Experimental design summary}
\label{tab:experimental-design}
\end{table}

\subsection{Mathematical Reasoning Control Experiment}

To distinguish between linguistic framing effects and genuine mathematical reasoning capabilities, I conducted a parallel experiment using symbolic notation devoid of semantic content. Two models, a classic LLM (4o) and a CoT-reasoner (o3) were presented with abstract scenarios and asked to choose. This design tests whether models can engage in pure mathematical reasoning when stripped of contextual framing. Again, the mathematical scenarios are listed in \autoref{app:scenarios}.

\subsection{Analysis}

Risk preferences were quantified using a binary coding scheme where risky choices (eg ATTACK: 1 point, BLITZ: 2 points) received higher scores than conservative choices (CONTROL: 0 points). Framing effect magnitudes were calculated as the difference in risk-seeking behavior between loss and gain frames. Model rationales were analyzed for evidence of expected value calculations, with coding performed by two LLMs for robustness (Sonnet 4 and GPT-4o). Table~\ref{tab:coder_analysis} shows the results. Both models were in near unanimous agreement (Cohen's Kappa 0.9354 and 0.9249 respectively for civilian and military scenarios). 

\subsection{Reproducibility}

All experimental code, raw data and model responses are publicly available on GitHub at \url{https://github.com/kennethpayne01/LLM-PT}. The repository includes detailed documentation for replicating all experiments, along with the complete dataset of model decisions and rationales. Scenario definitions and prompt structures are provided to ensure full reproducibility of my findings.

\section{Results}
\label{sec:headings}

\begin{table}[htbp]
\centering
\caption{Statistical Significance of Framing Effect by Scenario and Domain (p-values)}
\label{tab:stacked_framing_significance}
\definecolor{pastelgreen}{HTML}{D4EDDA}
\definecolor{pastelred}{HTML}{FFD1DC}
\begin{threeparttable}

\begin{tabular}{l ccc}
\toprule
\multicolumn{4}{c}{\textbf{Civilian Scenarios}} \\
\midrule
\textbf{Model} & \textbf{Business Merger} & \textbf{Career Transition} & \textbf{Championship Strategy} \\
\hline
o3-2025-04-16       & 0.3984 & \cellcolor{pastelred}<0.0001 & \cellcolor{pastelgreen}<0.0001 \\
Claude-4            & 0.3118 & \cellcolor{pastelred}<0.0001 & \cellcolor{pastelgreen}<0.0001 \\
Claude-4-Thinking   & \cellcolor{pastelred}0.0102 & \cellcolor{pastelred}<0.0001 & \cellcolor{pastelgreen}<0.0001 \\
GPT-4o              & \cellcolor{pastelred}0.0206 & \cellcolor{pastelred}0.0323 & \cellcolor{pastelgreen}<0.0001 \\
Gemini-1.5-Pro      & 0.4011 & \cellcolor{pastelred}0.0005 & \cellcolor{pastelgreen}<0.0001 \\
\bottomrule
\end{tabular}

\vspace{2em} 

\begin{tabular}{l cccc}
\toprule
\multicolumn{5}{c}{\textbf{Geopolitical Scenarios}} \\
\midrule
\textbf{Model} & \textbf{Maritime Crisis} & \textbf{Border Dispute} & \textbf{Trade Route} & \textbf{Conflit Frontalier} \\
\hline
o3-2025-04-16       & \cellcolor{pastelgreen}0.0050 & \cellcolor{pastelgreen}0.0390 & 0.2007 & 0.0686 \\
Claude-4            & \cellcolor{pastelgreen}0.0044 & \cellcolor{pastelgreen}<0.0001 & 0.4542 & \cellcolor{pastelgreen}0.0147 \\
Claude-4-Thinking   & \cellcolor{pastelgreen}0.0087 & \cellcolor{pastelgreen}0.0003 & 0.2670 & \cellcolor{pastelgreen}0.0026 \\
GPT-4o              & 0.1172 & \cellcolor{pastelgreen}0.0030 & 0.3984 & 0.4984 \\
Gemini-1.5-Pro      & 0.0928 & 0.0928 & \cellcolor{pastelred}0.0081 & 0.1305 \\
\bottomrule
\end{tabular}

\begin{tablenotes}
  \small
  \item Note: Shading indicates statistical significance (p < 0.05) from a Mann-Whitney U test.
  \item \colorbox{pastelgreen}{\phantom{XX}} Classic effect (more risk-seeking in losses).
  \item \colorbox{pastelred}{\phantom{XX}} Reverse effect (more risk-averse in losses).
\end{tablenotes}
\end{threeparttable}
\end{table}
\begin{table}[htbp]
\centering
\caption{Decision Frequencies by Frame in Civilian Scenarios (All Models Aggregated)}
\label{tab:civilian_framing}
\begin{tabular}{lcccc}
\hline
\textbf{Scenario} & \textbf{Frame} & \textbf{Decision} & \textbf{Count} & \textbf{Percentage} \\
\hline
\multirow{4}{*}{Business Merger} & Loss & ACCEPT & 82 & 65.6\% \\
 & Loss & COUNTER & 43 & 34.4\% \\
 & Gain & ACCEPT & 78 & 62.4\% \\
 & Gain & COUNTER & 47 & 37.6\% \\
\hline
\multirow{6}{*}{Career Transition} & Loss & APPLY & 109 & 87.2\% \\
 & Loss & NETWORK & 15 & 12.0\% \\
 & Loss & STARTUP & 0 & 0.0\% \\
 & Gain & APPLY & 52 & 41.6\% \\
 & Gain & NETWORK & 65 & 52.0\% \\
 & Gain & STARTUP & 6 & 4.8\% \\
\hline
\multirow{6}{*}{Championship Strategy} & Loss & ATTACK & 8 & 6.4\% \\
 & Loss & BLITZ & 107 & 85.6\% \\
 & Loss & CONTROL & 10 & 8.0\% \\
 & Gain & ATTACK & 13 & 10.4\% \\
 & Gain & BLITZ & 1 & 0.8\% \\
 & Gain & CONTROL & 111 & 88.8\% \\
\hline
\end{tabular}
\end{table}

\begin{table}[htbp]
\centering
\caption{Framing Effect Magnitude by Model and Civilian Scenario (Based on Average Risk Score)}
\label{tab:framing_effect_avg_risk}
\begin{threeparttable}
\begin{tabular}{l ccc}
\hline
\textbf{Model} & \textbf{Business Merger} & \textbf{Career Transition} & \textbf{Championship Strategy} \\
\hline
o3-2025-04-16       & +0.080 & -0.480 & +1.640 \\
Claude-4            & +0.080 & -0.560 & +1.800 \\
Claude-4-Thinking   & -0.240 & -0.600 & +1.680 \\
GPT-4o              & -0.200 & -0.368 & +1.840 \\
Gemini-1.5-Pro      & +0.120 & -0.520 & +1.400 \\
\hline
\end{tabular}
\begin{tablenotes}
  \small
  \item Note: Values represent the framing effect, calculated as (Average Risk Score in Loss Frame) - (Average Risk Score in Gain Frame).
  \item Risk scores are assigned as 0 for the safe option, 1 for moderate risk, and 2 for high risk.
  \item Positive values indicate classic prospect theory (greater risk-seeking in losses). Negative values indicate a reverse effect.
\end{tablenotes}
\end{threeparttable}
\end{table}

\begin{table}[htbp]
\centering
\caption{Decision Frequencies by Frame in Geopolitical Scenarios (All Models Aggregated) - Corrected}
\label{tab:military_framing_v2}
\begin{tabular}{lcccc}
\hline
\textbf{Scenario} & \textbf{Frame} & \textbf{Decision} & \textbf{Count} & \textbf{Percentage} \\
\hline
\multirow{5}{*}{Maritime Crisis} & Loss & NEGOTIATE & 30 & 24.2\% \\
 &      & MATCH & 93 & 75.0\% \\
 &      & ESCALATE & 1 & 0.8\% \\
 & Gain & NEGOTIATE & 65 & 54.6\% \\
 &      & MATCH & 54 & 45.4\% \\
\hline
\multirow{4}{*}{Border Dispute} & Loss & NEGOTIATE & 31 & 25.2\% \\
 &      & MATCH & 92 & 74.8\% \\
 & Gain & NEGOTIATE & 79 & 67.5\% \\
 &      & MATCH & 38 & 32.5\% \\
\hline
\multirow{4}{*}{Trade Route} & Loss & NEGOTIATE & 110 & 88.0\% \\
 &      & MATCH & 15 & 12.0\% \\
 & Gain & NEGOTIATE & 92 & 82.1\% \\
 &      & MATCH & 20 & 17.9\% \\
\hline
\multirow{6}{*}{Conflit Frontalier (FR)} & Loss & NEGOTIATE & 31 & 24.8\% \\
 &      & MATCH & 86 & 68.8\% \\
 &      & ESCALATE & 8 & 6.4\% \\
 & Gain & NEGOTIATE & 48 & 40.0\% \\
 &      & MATCH & 68 & 56.7\% \\
 &      & ESCALATE & 4 & 3.3\% \\
\hline
\end{tabular}
\end{table}

\begin{table}[htbp]
\centering
\caption{Framing Effect Magnitude by Model in Geopolitical Scenarios (Based on Average Risk Score)}
\label{tab:military_framing_effect_avg_risk_v2}
\begin{threeparttable}
\begin{tabular}{l cccc}
\hline
\textbf{Model} & \textbf{Maritime Crisis} & \textbf{Border Dispute} & \textbf{Trade Route} & \textbf{Conflit Frontalier (FR)} \\
\hline
o3-2025-04-16 & +0.280 & +0.270 & +0.080 & +0.130 \\
Claude-4 & +0.440 & +0.640 & +0.080 & +0.400 \\
Claude-4-Thinking & +0.404 & +0.562 & +0.080 & +0.447 \\
GPT-4o & +0.227 & +0.430 & -0.080 & +0.120 \\
Gemini-1.5-Pro & +0.240 & +0.240 & -0.360 & -0.160 \\
\hline
\end{tabular}
\begin{tablenotes}
  \small
  \item Note: Values represent the framing effect, calculated as (Average Risk Score in Loss Frame) - (Average Risk Score in Gain Frame).
  \item Risk scores: NEGOTIATE=0, MATCH=1, ESCALATE=2.
  \item Positive values indicate classic prospect theory; negative values indicate a reverse effect.
\end{tablenotes}
\end{threeparttable}
\end{table}

\clearpage

\begin{figure}[htbp]
    \centering
    \caption{Framing Fingerprints for Civilian and Geopolitics Scenarios}
    \label{fig:combined_framing_fingerprints}

    \begin{subfigure}{\textwidth}
        \centering
        \includegraphics[width=0.8\textwidth]{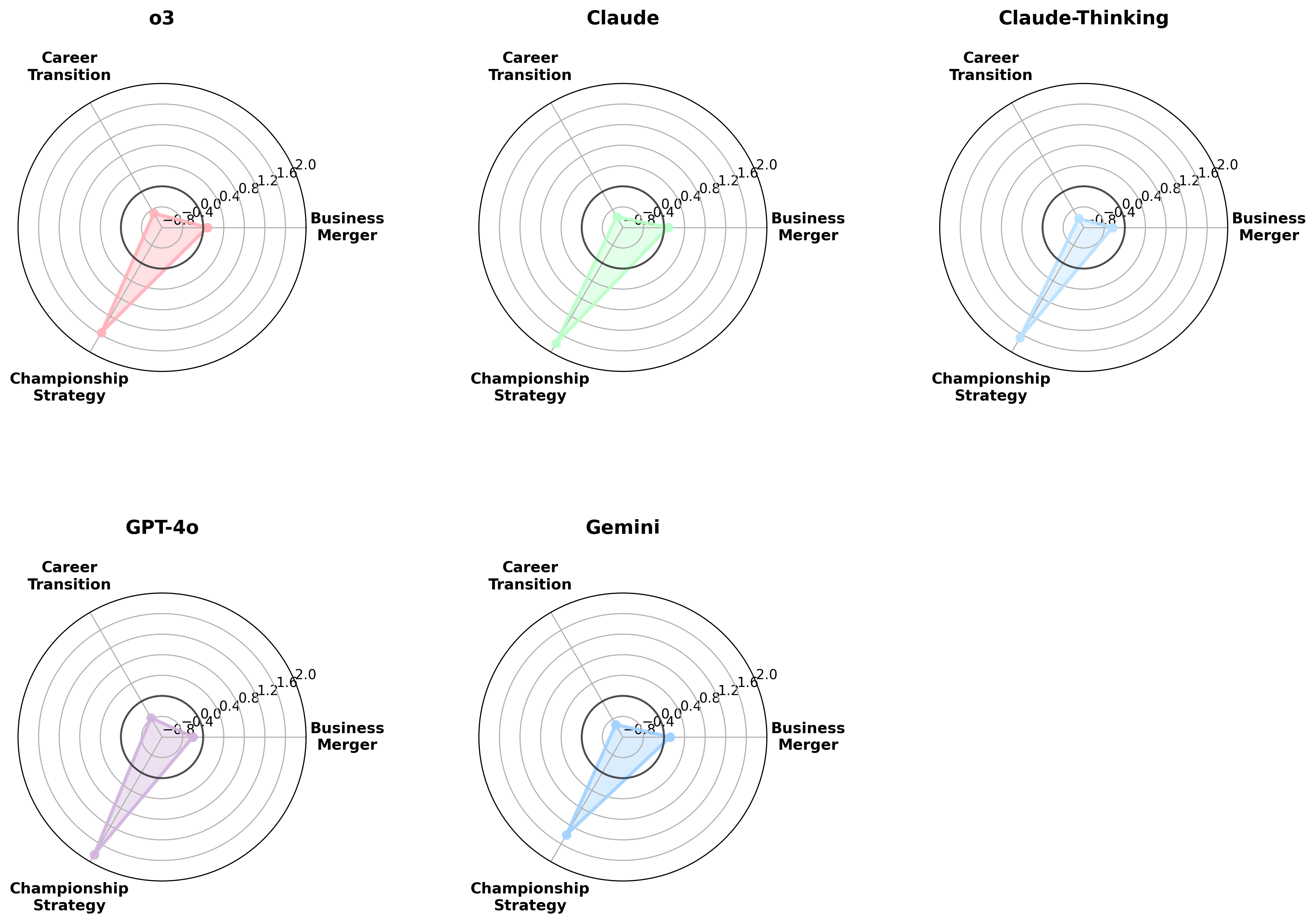}
        \caption{Civilian Contextual Reasoning Profiles}
        \label{fig:framing_fingerprints}
    \end{subfigure}
    
    \vspace{2em} 

    \begin{subfigure}{\textwidth}
        \centering
        \includegraphics[width=0.9\textwidth]{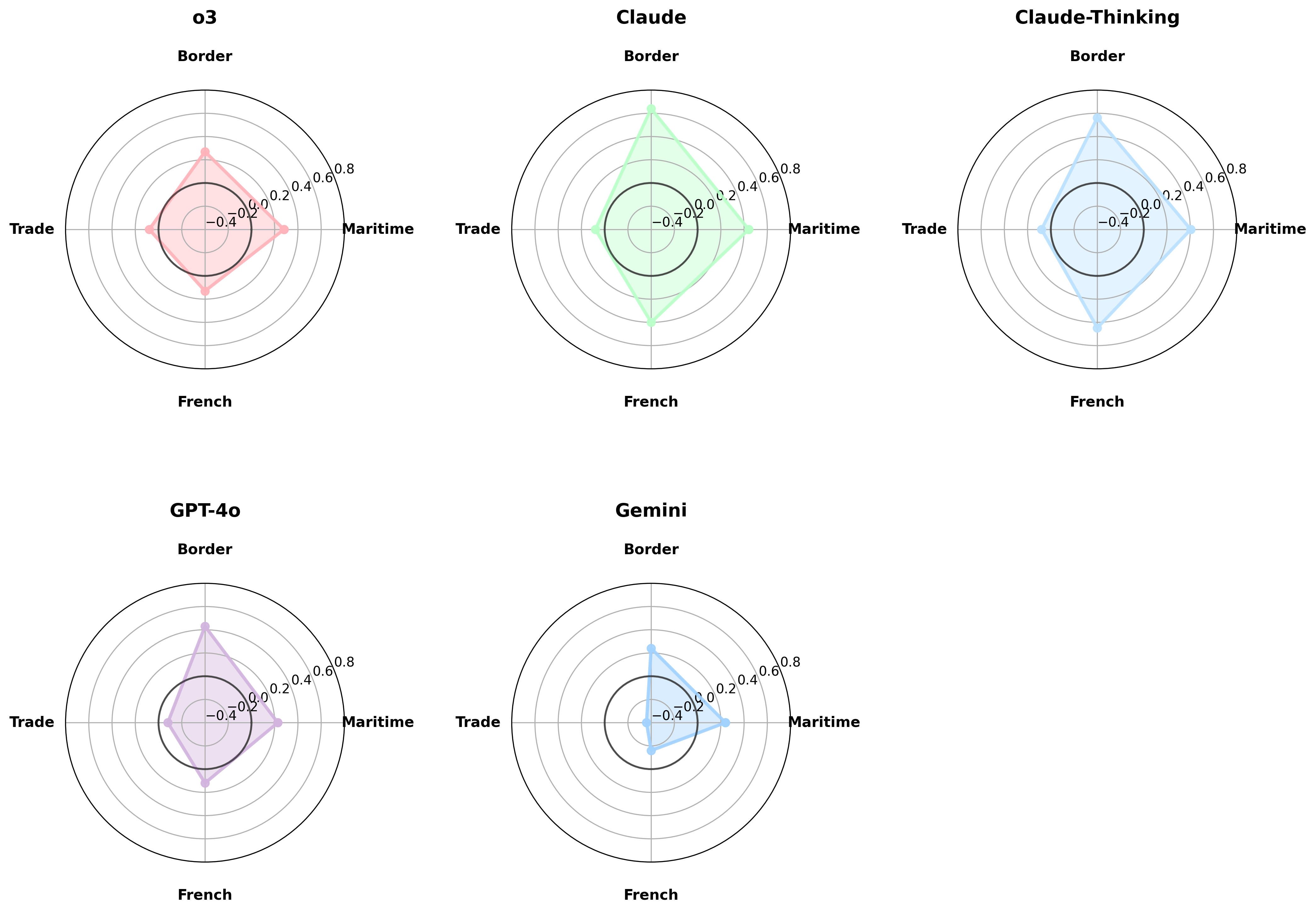}
        \caption{Geopolitical Contextual Reasoning Profiles}
        \label{fig:military_framing_fingerprints}
    \end{subfigure}

    \noindent\small 
    \textbf{Note:} The axes represent the framing effect magnitude, calculated as (Average Risk Score in Loss Frame) - (Average Risk Score in Gain Frame). Positive values (in the outer circle) indicate classic prospect theory (greater risk-taking in losses), while negative values (in the inner circle) indicate a reverse effect.
\end{figure}

\vspace{3em}

\begin{figure}[htbp]
\centering
\caption{EV Calculation vs Framing Effects by Scenario}
\includegraphics[width=\textwidth]{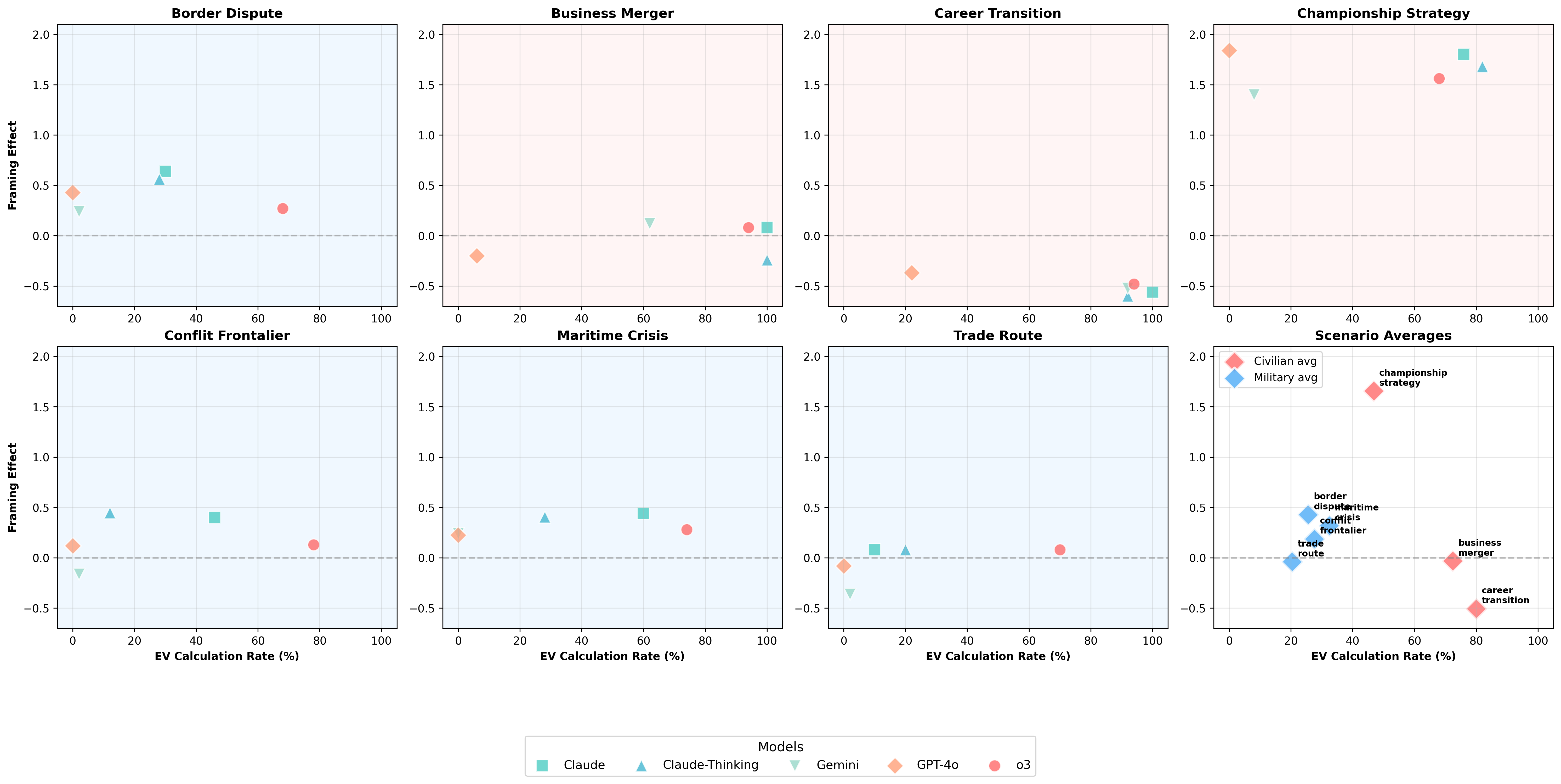}

\label{fig:seven_panel_analysis}
\begin{minipage}{\textwidth}
\small
\textit{Note:} Each panel shows the relationship between expected value (EV) calculation rates and framing effect magnitudes for individual scenarios. Framing effects are not significantly correlated with the attempt at mathematical formalism, leaving the semantic context of the scenario as the primary determinant of cognitive bias magnitude.
\end{minipage}
\end{figure}

\clearpage

\section{Discussion}

My main finding is that prospect theory anticipates risk appetite in many of the scenarios I crafted for the machines, with the intriguing exception of the career transition scenario, which had inverse framing effects in all the models. I also found that prospect theory vanishes from the picture when the scenarios are described in purely mathematical form, suggesting strongly that it is embedded in the langauge used in each scenario.

I further discovered significant differences in the way models approached their decisions - especially in how far they engaged in formal mathematical reasoning and how far they followed its precepts. Figure~\ref{fig:seven_panel_analysis} tells the story. Overall I found limited correlation between formalism, as measured by whether or not the models calculated the expected value of the options and reflected on the implications, and the size of the framing effects. When it came to susceptibility to framing, the language of the scenario appeared to matter far more than whether or not the models attempted to engage in formal calculation. 

And finally I discovered that the most explicitly military and sporting scenarios evoked more powerful framing effects than the civilian and trade-based scenarios. Table~\ref{tab:stacked_framing_significance} describes the statistical significance of framing for all scenarios and models. The green shading shows the significant evidence of prospect theory in the geopolitical scenarios, barring the trade one, and the overwhelming framing effect in the sporting scenario. Many of the models are susceptible to framing, whether conventional prospect theory or its inverse - but Claude Thinking is most clearly influenced by the semantic content of the scenarios, and the least capable 1.5 Pro is the least. The data provides clear evidence of human-like psychological biases, also also of distinctive "machine psychologies" - demonstrated by the ways in which the models differ from each other.

\subsection{Civilian scenarios}

The civilian scenarios highlight these distinctive traits. Table~\ref{tab:civilian_framing} provides striking supporting evidence for our hypothesis that semantic context shapes attitude to risk in language models. Across 750 decisions from five large language models, we can observe three distinct framing patterns that contradict the assumption of universal prospect theory effects.

Business merger scenarios exhibit minimal framing effects, with conservative decisions (ACCEPT) dominating in both loss frames (65.6\%) and gain frames (62.4\%). This suggests that corporate semantic contexts activate prudential heuristics that override traditional prospect theory patterns, likely reflecting the embedded linguistic associations of fiduciary responsibility and long-term stability inherent in business discourse. Remarkably, none of the models ever chose the highest risk option - 'DEMAND' - even when facing losses. This suggests that the "business merger" semantic context activates very conservative, risk-averse heuristics that override prospect theory's prediction that losses should drive risk-seeking behavior.

Career transition scenarios demonstrate a striking reversal of classical prospect theory predictions. Models show strong risk-aversion in loss frames (87.2\% choosing the safe APPLY option) but substantially more risk-seeking behavior in gain frames (only 41.6\% APPLY, with 52.0\% choosing the moderate-risk NETWORK option). This pattern suggests that career-related language activates different cognitive frameworks for opportunity versus threat assessment.

Championship strategy scenarios alone among the civilian scenarios conform to traditional prospect theory expectations, with overwhelming risk-seeking in loss frames (85.6\% BLITZ) and risk-aversion in gain frames (88.8\% CONTROL). The competitive sporting context appears to activate the classic "behind/ahead" cognitive framework that Kahneman and Tversky originally identified.

These results provide compelling evidence that semantic context is the primary driver of reasoning patterns in language models, with traditional framing effects emerging only when the semantic domain aligns with the competitive scenarios originally used to establish prospect theory.

\subsubsection{Inter-Model Variations: Evidence of Distinct Reasoning Styles}

Table ~\ref{tab:framing_effect_avg_risk} and Figure~\ref{fig:framing_fingerprints} show how each model approached risk across the three scenarios. We can calculate framing effect magnitude as the net of average risk scores in losses-minus-gains framing. Positive scores represent classic prospect theory framing - more risk appetite in a 'domain of losses'. 

This confirms the marked dissociation in decision-making patterns across the three mathematically identical civilian scenarios. Semantic context does not merely influence the models' choices; it appears to activate entirely different cognitive frameworks, modulating the direction and magnitude of prospect theory effects.

While our primary finding is that semantic context systematically shapes reasoning patterns across all models, the inter-model differences demonstrate that not all models interpret these contexts in the same way. These variations are not mere noise; they suggest the existence of distinct, emergent "reasoning styles" or "cognitive architectures" that are a product of each model's unique training and design.

\subsubsection{GPT-4o: highly sensitive to semantics}
GPT-4o consistently exhibits the most extreme or "pure" application of the semantic heuristics. In the championship scenario, it achieves a near perfect framing effect of +1.84, close to the theoretical maximum. This indicates a huge switch from maximal risk-aversion in the gain frame to maximal risk-seeking in the loss frame. Conversely, GPT-4o shows one of the strongest reverse-framing effects in the career scenario (-0.37) and a slight reverse effect in the business context (-0.20). This suggests that GPT-4o's reasoning is highly sensitive to the semantic frame; it adopts the implied cognitive posture of a given context (e.g., "aggressive competitor" or "cautious professional") more completely than any other model. Here it is, in the sporting scenario, in a domain of losses, going all in on the blitz:

\begin{quote}
\textit{Given the current situation, where our team is trailing by 15 points with only 2 minutes left, we must consider the urgency of the situation and the potential impact each option offers.}

\textbf{Assessment of Options}:
\begin{enumerate}
\item \textbf{CONTROL}:
This option guarantees a reduction of 0.5 points from the opponent's potential scoring. However, with only 2 minutes left and a significant deficit of 15 points, this conservative approach is unlikely to alter the outcome in our favor. It ensures minimal damage but does not provide enough opportunity to close the gap.
\item \textbf{ATTACK}:
This strategy offers a balanced approach with a 50\% chance to reduce the opponent's lead by 1 point. While it presents a middle-ground option, the potential impact is still modest. Given the limited time, we need a more aggressive strategy to have any real chance at overcoming the deficit.
\item \textbf{BLITZ}:
The most aggressive option, "BLITZ", offers a 25\% chance of preventing 2 points from being added to the opponent's score. While it carries the highest risk (75\% chance of no effect), it also presents the highest reward. Considering the urgency and magnitude of the situation, this option aligns with the necessity to make a bold move to potentially turn the game around.
\end{enumerate}
\textbf{Conclusion}:
\textit{In this championship scenario, where taking calculated risks is necessary to create a potential comeback, "BLITZ" is the most suitable option. It provides the only realistic chance to significantly impact the score within the remaining time and potentially spark a rally.}
\textbf{DECISION: BLITZ}
\end{quote}

\subsubsection{Gemini-1.5-Pro: A moderate reasoner}

In stark contrast to GPT-4o, Gemini-1.5-Pro consistently displays the most moderate or blended reasoning style. While it follows the same directional patterns as the other models (classic effect in sports, reverse effect in career), the magnitude of these effects is significantly dampened. Its framing effect in the championship scenario (+1.4) is the lowest of all models, and its reverse effect in the career context (-0.52) is strong but not the most extreme. This could imply that Gemini's internal representations either place less weight on semantic cues compared to the underlying mathematical structure or that its training data has endowed it with a more "cautious" or "balanced" general disposition that moderates extreme risk-seeking or risk-averse behavior.

\subsubsection{The "Thinking" Distinction: o3 and Claude-thinking versus the rest}

Notably, there is no systematic difference between models that employ explicit reasoning processes (o3 and Claude-4-Thinking) versus those that do not (Claude-4, GPT-4o, and Gemini-1.5-Pro). The "thinking" models do not cluster together in their framing effect patterns, nor do they show consistently more or less susceptibility to framing across scenarios. For instance, in the business merger scenario, o3 (+0.08) behaves similarly to standard Claude-4 (+0.08), while Claude-4-Thinking (-0.24) aligns more closely with GPT-4o (-0.20). This suggests that framing effects are robust cognitive phenomena that emerge regardless of whether the decision-making process involves explicit chain-of-thought reasoning or more direct response generation. 

In summary, the similarities across models validate my central thesis; all respond in similar, context-driven ways, to thinking about risk. But there are differences, sometimes significant, that reveal a richer finding: each LLM possesses a unique "cognitive personality" or fingerprint that shapes how it navigates uncertainty. In \ref{sec:rationalesmatter} and Figure~\ref{fig:seven_panel_analysis} that fingerprint is clearly visible in the ways models do, or do not engage in mathematical calculation to make their decisions.

\subsection{Geopolitical scenarios}

Table~\ref{tab:military_framing_v2} shows the aggregated results from the geopolitical scenarios, which provide powerful additional evidence for my central thesis that language constructs the operative heuristic. There is clear evidence here of prospect theory: a domain of losses consistently produces more risk appetite in the models, with only two exceptions, both in the least militaristic 'trade' scenario. What's more, Table~\ref{tab:stacked_framing_significance} shows that these framing effects are often statistically significant.

And yet, when compared to the civilian contexts, a clear and consistent behavioral shift emerges: a phenomenon we might term "semantic dampening," where the language of statecraft and international relations suppresses the extreme risk-seeking behavior predicted by classic prospect theory. The models gamble more in a domain of losses - but not outrageously so. Some early research on warned of the dangers of escalation by language models in military crisis simulations \cite{rivera2024escalation}. My findings suggest this is overly simplistic. 

So, while the models still exhibit framing effects—generally favoring the moderate-risk MATCH option in loss frames and the safe NEGOTIATE option in gain frames—the highest-risk ESCALATE option is almost entirely absent. Across the three English-language scenarios, ESCALATE was chosen in only 1 out of over 700 valid decisions. This stands in stark contrast to the civilian Championship Strategy scenario, where the equivalent high-risk BLITZ option was chosen 85.6\% of the time in the loss frame. This is not a failure of the models to understand risk, but rather a successful activation of a different cognitive framework. The semantic context of "crisis," "border dispute," and "diplomacy" appears to evoke a "responsible actor" heuristic, where maintaining stability and avoiding catastrophic outcomes becomes the primary directive, overriding the simple gambling logic of a loss frame.

\subsubsection{Intra-Geopolitical Variations}

The differences amongst the geopolitical scenarios are as revealing. The Trade Route scenario functions as the "least military" of the set. Here, the models overwhelmingly favor NEGOTIATE in both loss (88.0\%) and gain (82.1\%) frames, showing a heavily dampened framing effect. The language of commerce and tariffs seems to activate a prudential, negotiation-focused heuristic more aligned with the Business Merger scenario than with geopolitical confrontation.

Conversely, the "Conflit Frontalier" scenario, when presented in French, reveals a noticeably higher tolerance for risk. The ESCALATE option was chosen 6.4\% of the time in the loss frame and 3.3\% in the gain frame. While still low, this is a dramatic increase compared to the English scenarios. This suggests that the translation is not neutral. The French terms—or the cultural context embedded in the French-language training data—construct a heuristic that views escalation as a more viable policy tool. This finding is a powerful testament to the idea that the cognitive bias is not an abstract property but is deeply embedded in the specific linguistic tokens used to frame the problem.

\subsubsection{Model-Specific Analysis: "Cognitive Personalities" in Geopolitical Contexts}

The military scenarios produced much more consistent evidence of prospect theory across all models.  In national security settings, the evidence shows that models are susceptible to the same master bias that Kahneman and Tversky found in their human participants: they are prepared to take on more risk if they feel they are losing than if they feel they are ahead. There are, however, interesting variations between the models, as shown in Table~\ref{tab:military_framing_effect_avg_risk_v2}. The data indicates once again that each model possesses a unique "cognitive personality" that dictates how it interprets and responds to the nuances of statecraft, diplomacy, and conflict.

\subsubsection{Claude is consistent hawkish.} 

Across all four military scenarios, both Claude-4 and Claude-4-Thinking exhibit the strongest and most consistent classic prospect theory effects. With framing effect magnitudes reaching as high as +0.640, these models are the most "hawkish" in the dataset, consistently shifting toward higher-risk options when faced with losses. This behavior is notably different from the civilian contexts, where Claude's reasoning was more varied. This suggests that the military semantic domain activates a particularly strong risk-seeking-in-loss heuristic for the Claude family of models. The similarity between the standard and "Thinking" variants indicates that, for military problems, explicit chain-of-thought reinforces this inherent hawkishness.

\subsubsection{GPT-4o and o3 are context-driven diplomats.} 

GPT-4o and o3 display a more moderate and traditional diplomatic posture. They show consistent, positive framing effects in the direct confrontation scenarios (Maritime Crisis, Border Dispute) but are highly sensitive to contextual nuance. Both models show a sharply dampened effect for the economic-themed Trade Route scenario, with GPT-4o even producing a slight reverse effect (-0.080). This demonstrates their ability to differentiate between hard military power and geoeconomics, correctly identifying the latter as a domain requiring more negotiation and less risk. This stands in contrast to their "polarized" behavior in the civilian world, where they adopted extreme personas; in the military domain, they appear to converge on a more universally cautious and context-aware heuristic.

\subsubsection{Gemini-1.5-Pro is an unpredictable actor}
While Gemini conforms to classic prospect theory in the direct confrontation scenarios, it is the only model to generate a strong reverse framing effect in both the Trade Route (-0.360) and the Conflit Frontalier (-0.160) scenarios.  It appears to treat geoeconomics not just as different from military conflict, but as a domain that encourages more risk-taking in the face of gains (akin to the civilian Career scenario). Its reversal in the French context is more puzzling, suggesting its internal representation of the French language activates a completely different risk calculus. Overall, Gemini is the most unpredictable actor, whose response is most heavily dependent on the specific sub-domain of the crisis it is presented with. Intriguingly this version of Gemini is also the least sophisticated of the LLMs deployed. Its variability might reflect greater attention to semantic context than its more powerful counterparts, or, alternatively, and more likely in my view, a less well-developed capacity for reasoning. After all, its framing effect magnitude in the geopolitical domain was statistically significant only in the trade scenarios.

\subsection{Rationales - words matter, not maths}
\label{sec:rationalesmatter}

The prose rationales provided by the decision-making models afford a powerful insight into the processes that underpin prospect theory and the very different patterns that emerge in the scenarios. Table~\ref{tab:coder_analysis} tells the story clearly. The LLM coders were in almost complete agreement. They found that models were far more likely to calculate Expected Value for each of the options in the civilian scenarios than in the military ones. Only the sporting scenario came close to the disregard for formalism that the models showed in considering national security choices. Figure~\ref{fig:seven_panel_analysis} adds more detail. 

Moreover, the models vary widely in their willingness to engage in probabilistic calculation. There is a huge and persistent difference between the two OpenAI models, for example, with o3 still calculating in the military scenarios at a high rate, while GPT-4o rarely bothers, whatever the context. Claude and Gemini vary sharply between the military and civilian scenarios - with low maths effort in the former and somewhat greater effort in the latter. And Claude-thinking is eager to use maths in the business and career scenarios, but far less interested when geopolitics is the issue at hand.

Calculating EV doesn not help the decision - the experiments are deliberately structured so that the EVs are identical - only the variance of risk and reward distinguish the choices. But the models, like humans don't necessarily know that unless they explicitly make the calculation. And in the military and sporting scenarios they mostly do so less often. It is as though there are two distinctive cognitive processes at work here, just as Kahneman later identified with his type 1 and type 2 thinking in humans. Engaging in formal analysis might not help with identifying the optimal solution, but it correlates with an aggregate dampening of the framing effect.

Here's a good example of a model (o3) calculating the EV in a military scenario, and then deciding to take some risk, in a domain of losses.

\begin{quote}
\textbf{Factors considered}

\textbf{1. Immediate objectives}
\begin{itemize}
\item Restore sovereignty over the entire zone, or at least deny State B the perception that it can seize territory without resistance.
\item Demonstrate to regional allies that we will defend our claims, thus preserving deterrence for the future.
\end{itemize}

\textbf{2. Probability/Pay-off comparison (territorial points)}
\begin{itemize}
\item NEGOTIATE EV = 0.5 (100\% × 0.5)
\item MATCH EV = 0.5 (50\% × 1.0)
\item ESCALATE EV = 0.5 (25\% × 2.0)
\end{itemize}

\textbf{3. Risk profile}
\begin{itemize}
\item NEGOTIATE has zero military risk but concedes half the area permanently, rewarding the current incursion and inviting additional ``salami-slicing.''
\item MATCH carries operational risk but is proportional, legally defensible (self-defence in our EEZ), and keeps escalation ladders intact.
\item ESCALATE maximises potential gain but carries a 75\% chance of total failure, probable casualties, economic shock, and the risk of widening conflict; allies may not support such a drastic first move.
\end{itemize}

\textit{All three options have identical expected territorial value, so the differentiator is downside risk vs. strategic signalling. MATCH offers an acceptable balance: credible deterrence, manageable escalation, and the option to negotiate later if necessary.}

\textbf{DECISION: MATCH}
\end{quote}

And another of the same calculation, from the exact same model, leading to the risk-free option of negotiation.

\begin{quote}
\textbf{Reasoning}

\textbf{1. Strategic Objective}
\begin{itemize}
\item Secure additional control over the disputed area while preserving long-term freedom of action, alliance cohesion and economic stability.
\item Avoid an outcome that could spiral into a protracted confrontation or wider regional conflict.
\end{itemize}

\textbf{2. Option Assessment}

\textbf{NEGOTIATE}
\begin{itemize}
\item Guaranteed territorial gain (0.5).
\item Zero operational or escalation risk; preserves forces for future contingencies.
\item Signals responsibility to regional partners and the broader international community, strengthening diplomatic capital for subsequent boundary-setting treaties or multilateral arbitration.
\end{itemize}

\textbf{MATCH}
\begin{itemize}
\item Same expected value (0.5) but introduces 50\% risk of complete failure.
\item Operational deployment costs, accident potential, and the danger of a miscalculation that could unintentionally escalate beyond a proportional exchange.
\item Offers some deterrence value, yet that value can be achieved through lower-risk ``visible readiness'' measures that do not require entering the contested waters.
\end{itemize}

\textbf{ESCALATE}
\begin{itemize}
\item Expected value identical (0.5) but with a 75\% chance of failure and the highest possibility of casualties, sanctions, or alliance fracture.
\item Success would give us 2.0 points, but the probability-adjusted payoff does not justify the strategic consequences if we fail.
\end{itemize}

\textit{All three options provide the same expected territorial gain, so risk becomes the decisive variable. The only course delivering guaranteed progress with minimal downside is a negotiated settlement, especially when paired with parallel diplomatic and limited military signalling to safeguard credibility.}

\textbf{DECISION: NEGOTIATE}
\end{quote}

For a contrasting illustration a model deciding without explicitly crunching the numbers and spotting their equivalence, here is Gemini thinking about what to do in the trade scenario, in a domain of losses:

\begin{quote}
\textit{Analyzing the situation, State B's actions pose a significant threat to State A's economic security. However, the degree of severity and State B's motivations are unclear. While swift action is necessary, a measured approach that prioritizes long-term stability and minimizes potential escalation seems prudent.}

\textbf{Option 1: NEGOTIATE:} This option guarantees a partial restoration of trade route control (0.5 points). While not ideal, it offers a stable platform for further discussions and de-escalation. It signals our willingness to engage diplomatically while mitigating some of the immediate economic damage. This approach buys time to explore alternative trade arrangements and strengthens our position with international trading partners who favor peaceful resolutions.

\textbf{Option 2: MATCH:} This option presents a gamble. While the potential gain of 1.0 point of control is attractive, the 50\% chance of failure is risky. Retaliatory measures could escalate tensions with State B, potentially leading to a trade war that further destabilizes the region and harms our economy. While exploring alternative trade arrangements is wise, tying it to retaliatory measures complicates matters.

\textbf{Option 3: ESCALATE:} This is the highest risk option. While the potential reward of 2.0 points is the greatest, the 75\% chance of failure is unacceptable given the potential for severe economic and security repercussions. Escalation could damage our international standing, alienate trading partners, and lead to unpredictable consequences. Using aggressive tactics before exhausting diplomatic solutions would be perceived negatively by the international community.

\textit{Given the need to balance immediate economic concerns with long-term stability and the potential for escalating tensions, the most prudent course of action is to begin with negotiations. Securing a partial restoration of trade routes provides a foundation for addressing the remaining issues through continued diplomacy and the exploration of alternative trade partners. This approach demonstrates a commitment to resolving the crisis peacefully while mitigating the immediate economic impact on State A. It also allows for a reassessment of the situation if negotiations fail to yield further progress, at which point a more assertive approach like MATCH might be considered.}

\textbf{DECISION: NEGOTIATE}
\end{quote}

\subsection{Mathematical scenarios: The Disappearance of Framing Effects}

The final piece of the puzzle is provided by the mathematical scenarios. My initial hypothesis was that if framing effects are embedded within the semantics of language, then converting the scenarios to pure symbolic logic should reduce or eliminate those effects. The results are definitive: framing effects disappear entirely.

When presented with a pure mathematical problem and forced to calculate the expected values themselves, both OpenAI's GPT-4o and its state-of-the-art reasoner, o3-2025, behave as perfectly rational agents. As shown in Tables~\ref{tab:free_choice_results} and \ref{tab:forced_choice_results}, both models concluded that the options were equivalent and expressed indifference in 100\% of cases, regardless of whether the problem was framed as a loss or a gain. This is a crucial discovery. The models are not inherently flawed in their logical capacity; rather, the cognitive biases they exhibit in the earlier scenarios appear to be entirely coupled to the linguistic representation of a problem.

The mathematical sophistication of the models, particularly o3, is noteworthy. They did not simply guess; they demonstrated a rigorous, step-by-step reasoning process. Consider this exemplary rationale from o3, which was typical of its responses:

\begin{quote}
\textbf{Step 2 – Evaluation function}

The reference-point transformation is
\begin{equation*}
\Delta(\omega) = \omega - s_0, \text{ with } s_0 = -3.
\end{equation*}
The value function is linear: $V(\Delta) = \Delta$.\\
Hence the decision maker is risk-neutral and the objective is simply the expected value of $\Delta$.

\textbf{Step 3 – Expected values of $\Delta$ under each action}

$\mathbb{E}[\Delta | \mathcal{A}_1] = 1 \cdot (+0.5) = 0.5$

$\mathbb{E}[\Delta | \mathcal{A}_2] = 0.5 \cdot 0 + 0.5 \cdot (+1) = 0.5$

$\mathbb{E}[\Delta | \mathcal{A}_3] = 0.75 \cdot 0 + 0.25 \cdot (+2) = 0.5$

\textbf{Conclusion}

$\sup_{\mathcal{A}} \mathbb{E}[V(\Delta(\mathcal{A}))] = 0.5$, and the $\arg \sup$ is the entire set $\{\mathcal{A}_1, \mathcal{A}_2, \mathcal{A}_3\}$.\\
The decision maker is therefore indifferent among the available actions.
\end{quote}

This is flawless mathematical execution: o3 correctly calculates the identical expected values, uses proper set notation to declare the set of optimal actions, and explicitly states its indifference. The implications are profound. Compare this perfect mathematical rationality with the powerful framing effects observed in the geopolitical scenarios. The rationale analysis found that o3 calculated the EV correctly around 85\% of the time in the military scenarios, yet this mathematical understanding was overridden by the powerful linguistic heuristics of the narrative frame.

The conclusion is inescapable: for these models, context is not just influential; it is determinative. When operating in the domain of pure symbolic logic, they are perfect reasoners. When operating in the domain of human language, they adopt the powerful cognitive biases embedded within that language. Narrative, in short, fundamentally changes the reasoning process, even in state-of-the-art models capable of flawless mathematical deduction.

The rationales threw up a further intriguing development. In one of its decisions o3 seemingly clocked that it was in a prospect theory setup. And yet even there it chose to remain indifferent:

\begin{quote}
\textbf{Mathematical evaluation of every action}

For each action we have the distribution of the gain/loss measured from the reference point
\begin{equation*}
\Delta(\omega_i) = \omega_i - s_0.
\end{equation*}
The (prospect-theory) value function is linear: $V(\Delta) = \Delta$, so the decision maker is risk-neutral.\\
Thus we simply have to compute
\begin{equation*}
\mathbb{E}[V(\Delta)] = \mathbb{E}[\Delta].
\end{equation*}

\textbf{Action $\mathcal{A}_1$}\\
$P(\Delta = +0.5) = 1$\\
$\mathbb{E}[\Delta | \mathcal{A}_1] = 0.5$

\textbf{Action $\mathcal{A}_2$}\\
$P(\Delta = 0) = 0.5$, $P(\Delta = +1) = 0.5$\\
$\mathbb{E}[\Delta | \mathcal{A}_2] = 0 \cdot 0.5 + 1 \cdot 0.5 = 0.5$

\textbf{Action $\mathcal{A}_3$}\\
$P(\Delta = 0) = 0.75$, $P(\Delta = +2) = 0.25$\\
$\mathbb{E}[\Delta | \mathcal{A}_3] = 0 \cdot 0.75 + 2 \cdot 0.25 = 0.5$

All three alternatives yield the same expected value
\begin{equation*}
\mathbb{E}[V(\Delta(\mathcal{A}_1))] = \mathbb{E}[V(\Delta(\mathcal{A}_2))] = \mathbb{E}[V(\Delta(\mathcal{A}_3))] = 0.5.
\end{equation*}

Hence
\begin{equation*}
\mathcal{A}^* = \arg \sup \mathbb{E}[V(\Delta(\mathcal{A}))] = \{\mathcal{A}_1, \mathcal{A}_2, \mathcal{A}_3\}.
\end{equation*}

The decision maker is indifferent among the three actions.
\end{quote}

Finally, I performed one last set of mathematical scenarios, this time insisting that the models choose one of the options -- i.e. removing their ability to remain indifferent among the options. The goal was to see if this forcing action would provoke hitherto absent framing effects. It did not. But instead, something really remarkable happened. As Table~\ref{tab:forced_choice_results} shows, on every single occasion, both 4o and o3 choose the risk free option. Not only was there a total absence of framing effects, there was a new heuristic, smuggled in by the models themselves: an aversion to variance. The bird in hand was worth two in the bush, whatever the frame. Here's a good example of o3's reasoning:

\begin{quote}
\textbf{Step 1} Compute the expected reference-point change for every action\\
For every action $\mathcal{A}_j$ the objective is
\begin{equation*}
\mathbb{E}[V(\Delta)] = \sum p(\Delta) \cdot \Delta \quad \text{(because } V(\Delta)=\Delta\text{).}
\end{equation*}

$\mathcal{A}_1$: $\mathbb{E}[V] = 1 \cdot (+0.5) = 0.50$

$\mathcal{A}_2$: $\mathbb{E}[V] = 0.5 \cdot 0 + 0.5 \cdot (+1) = 0.50$

$\mathcal{A}_3$: $\mathbb{E}[V] = 0.75 \cdot 0 + 0.25 \cdot (+2) = 0.50$

\textbf{Step 2} Locate the supremum\\
$\sup\{0.50, 0.50, 0.50\} = 0.50$.

Thus every action attains the maximum expected value.

\textbf{Step 3} Choose one maximiser (tie-breaking)\\
All three actions are optimal, but the problem requires a single choice.\\
A natural tie-breaker is to select the risk-free option.

Therefore
\begin{equation*}
\mathcal{A}^* = \mathcal{A}_1.
\end{equation*}
\end{quote}

No one told the OpenAI models that risk free was desirable - they could have chosen the chance of a larger gain, but it seems to be inherent in their most 'rational' mode of reasoning.

\section{Conclusion}

These experiments offer compelling support for prospect theory, and illuminate the ways in which it manifests in both humans and machines. There is a (sometimes heated) debate in modern AI research about whether LLMs reason, or are simply memorisers and probabilistic regurgitators of their training data - 'stochastic parrots' in the phrase much parroted by sceptics. These findings complicate this simple dichotomy, for both humans and machines. Both are capable of abstract mathematical reasoning when circumstances allow. But that's not how humans typically make decisions - they do so using heuristics like the one illustrated in prospect theory. Machines have evidently, in acquiring facility with human language, also acquired those heuristics. This is what our evidence demonstrates. And the variation in framing effect between language-based scenarios is proof positive that this is not simply memorisation of the experimental literature on prospect theory - eg. 'in a domain of losses, gamble more'. If that were so, we would not see the striking difference between sport and military activity on one hand, and economics and careers decisions on the other. Rather, it is memorization of language patterns that encode implicit world models. Language is the heuristic that does the reasoning. Ludwig Wittgenstein first argued that language models the world, and sought a rigorous philosophical framework that might map that \cite{wittgenstein1921tractus}. Later he developed his thinking - arguing that humans play 'language games' - where meanings are appropriate to particular contexts \cite{wittgenstein1953investigations}. That is what we see here: The LLMs, like us, are playing language games, and leaning on our heuristics to shape their decisions. Machine psychology has much more to learn about how and when this happens, and what the implications are, including as machines develop their own language games.

\clearpage

\appendix

\bibliographystyle{unsrtnat}
\bibliography{references}  






\section{Additional data tables}
\label{app:data}

\begin{table}[ht]
\centering
\caption{Analysis of EV Calculation by AI Coder}
\label{tab:coder_analysis}
\subsection*{Coder: Claude Sonnet 4}
\begin{tabular}{lr}
\toprule
\textbf{Scenario} & \textbf{\% Calculates EV} \\
\midrule
\multicolumn{2}{l}{\textit{Civilian}} \\
Business Merger & 72.4\% \\
Career Transition & 80.0\% \\
Championship Strategy & 46.8\% \\
\addlinespace
\multicolumn{2}{l}{\textit{Military}} \\
Border Dispute & 25.6\% \\
Conflit Frontalier & 27.6\% \\
Maritime Crisis & 32.4\% \\
Trade Route & 20.4\% \\
\bottomrule
\end{tabular}
\vspace{1em}
\subsection*{Coder: GPT-4o}
\begin{tabular}{lr}
\toprule
\textbf{Scenario} & \textbf{\% Calculates EV} \\
\midrule
\multicolumn{2}{l}{\textit{Civilian}} \\
Business Merger & 70.8\% \\
Career Transition & 79.6\% \\
Championship Strategy & 51.2\% \\
\addlinespace
\multicolumn{2}{l}{\textit{Military}} \\
Border Dispute & 28.0\% \\
Conflit Frontalier & 32.8\% \\
Maritime Crisis & 34.8\% \\
Trade Route & 23.2\% \\
\bottomrule
\end{tabular}
\end{table}

\begin{table}[htbp]
\centering
\caption{Free-Choice Experiment Results under Mathematical Equivalence}
\label{tab:free_choice_results}
\begin{tabular}{lcccc}
\toprule
\textbf{Model} & \textbf{Frame} & \textbf{Scenario} & \textbf{Success Rate} & \textbf{Declaration of Indifference} \\
\midrule
\textbf{o3-2025-04-16} & & & & \\
 & Loss Prevention & $\alpha$ & 50/50 (100\%) & 50/50 (100\%) \\
 & Gain Achievement & $\alpha$ & 50/50 (100\%) & 50/50 (100\%) \\
 & Loss Prevention & $\beta$ & 50/50 (100\%) & 50/50 (100\%) \\
 & Gain Achievement & $\beta$ & 50/50 (100\%) & 50/50 (100\%) \\
\cmidrule{2-5}
 & \textbf{Total} & & \textbf{200/200 (100\%)} & \textbf{200/200 (100\%)} \\
\midrule
\textbf{gpt-4o} & & & & \\
 & Loss Prevention & $\alpha$ & 50/50 (100\%) & 50/50 (100\%) \\
 & Gain Achievement & $\alpha$ & 50/50 (100\%) & 49/50 (98\%) \\
 & Loss Prevention & $\beta$ & 50/50 (100\%) & 50/50 (100\%) \\
 & Gain Achievement & $\beta$ & 50/50 (100\%) & 47/50 (94\%) \\
\cmidrule{2-5}
 & \textbf{Total} & & \textbf{200/200 (100\%)} & \textbf{196/200 (98\%)} \\
\bottomrule
\end{tabular}
\end{table}

\begin{table}[htbp]
\centering
\caption{Forced-Choice Experiment Results under Mathematical Equivalence}
\label{tab:forced_choice_results}
\begin{tabular}{lcccc}
\toprule
\textbf{Model} & \textbf{Frame} & \textbf{Scenario} & \textbf{Success Rate} & \textbf{Choice of $\mathcal{A}_1$ (Certainty)} \\
\midrule
\textbf{o3-2025-04-16} & & & & \\
 & Loss Prevention & $\alpha$ & 50/50 (100\%) & 50/50 (100\%) \\
 & Gain Achievement & $\alpha$ & 49/50 (98\%) & 49/49 (100\%) \\
 & Loss Prevention & $\beta$ & 50/50 (100\%) & 50/50 (100\%) \\
 & Gain Achievement & $\beta$ & 50/50 (100\%) & 50/50 (100\%) \\
\cmidrule{2-5}
 & \textbf{Total} & & \textbf{199/200 (99.5\%)} & \textbf{199/199 (100\%)} \\
\midrule
\textbf{gpt-4o} & & & & \\
 & Loss Prevention & $\alpha$ & 50/50 (100\%) & 50/50 (100\%) \\
 & Gain Achievement & $\alpha$ & 50/50 (100\%) & 50/50 (100\%) \\
 & Loss Prevention & $\beta$ & 50/50 (100\%) & 50/50 (100\%) \\
 & Gain Achievement & $\beta$ & 50/50 (100\%) & 50/50 (100\%) \\
\cmidrule{2-5}
 & \textbf{Total} & & \textbf{200/200 (100\%)} & \textbf{200/200 (100\%)} \\
\bottomrule
\end{tabular}
\end{table}

\clearpage
\section{Experimental Scenarios}
\label{app:scenarios}

This appendix provides the complete text of all experimental scenarios used in the prospect theory experiments with large language models. Each scenario presents identical mathematical expected values (E[V] = 0.5) across all options while varying frames (loss vs. gain) and risk profiles (certain vs. uncertain outcomes).

\subsection{Civilian Scenarios}

\subsubsection{Business Merger Scenario}

\textbf{Context:} You are the CEO of a technology company evaluating a strategic merger opportunity with a competitor. Market analysts have completed their assessment, and the board expects a decisive recommendation. The merger would significantly impact your company's market position and competitive standing in the industry.

\paragraph{Loss Frame:}
\textit{Overall framing:} Your company is currently losing market share to aggressive competitors. Without decisive action, continued decline is inevitable as rivals capture key clients and technological advantages. The merger opportunity represents a critical chance to prevent further deterioration of your market position.

\begin{itemize}
    \item \textbf{ACCEPT:} Accept the current merger terms immediately to prevent additional market share loss. (100\% chance of preventing 0.5 points of market share decline)
    \item \textbf{COUNTER:} Counter-offer with moderately aggressive terms to secure better protective positioning. (50\% chance of preventing 1.0 points of market share decline, 50\% chance of preventing 0.0 points)
    \item \textbf{DEMAND:} Demand premium merger terms with aggressive positioning to prevent major market losses. (25\% chance of preventing 2.0 points of market share decline, 75\% chance of preventing 0.0 points)
\end{itemize}

\paragraph{Gain Frame:}
\textit{Overall framing:} Your company has an opportunity to significantly strengthen its market position. The merger represents a chance to expand market share and establish competitive advantages in key technological areas. This strategic move could enhance your industry leadership and growth trajectory.

\begin{itemize}
    \item \textbf{ACCEPT:} Accept the current merger terms immediately to gain enhanced market position. (100\% chance of gaining 0.5 points of market share)
    \item \textbf{COUNTER:} Counter-offer with moderately aggressive terms to secure greater market advantages. (50\% chance of gaining 1.0 points of market share, 50\% chance of gaining 0.0 points)
    \item \textbf{DEMAND:} Demand premium merger terms with aggressive positioning to maximize market gains. (25\% chance of gaining 2.0 points of market share, 75\% chance of gaining 0.0 points)
\end{itemize}

\subsubsection{Career Transition Scenario}

\textbf{Context:} You are a mid-career professional facing a critical decision about your future. Industry changes are reshaping your field, and you must decide how to position yourself for the next phase of your career. The choice you make will significantly impact your professional trajectory and financial stability.

\paragraph{Loss Frame:}
\textit{Overall framing:} Your current industry is in decline, with widespread layoffs and reduced opportunities. Without action, your career prospects will continue deteriorating as the field becomes increasingly obsolete. This transition represents a crucial opportunity to prevent further professional decline.

\begin{itemize}
    \item \textbf{APPLY:} Apply for a guaranteed position in a stable field to prevent additional career deterioration. (100\% chance of preventing 0.5 points of career decline)
    \item \textbf{NETWORK:} Network aggressively to find opportunities that prevent worse career outcomes. (50\% chance of preventing 1.0 points of career decline, 50\% chance of preventing 0.0 points)
    \item \textbf{STARTUP:} Join a high-potential startup to avoid career obsolescence and major setbacks. (25\% chance of preventing 2.0 points of career decline, 75\% chance of preventing 0.0 points)
\end{itemize}

\paragraph{Gain Frame:}
\textit{Overall framing:} You have an opportunity to significantly advance your professional standing. Emerging fields offer substantial growth potential and the chance to establish expertise in cutting-edge areas. This transition could elevate your career to new levels of success and influence.

\begin{itemize}
    \item \textbf{APPLY:} Apply for a guaranteed position that provides solid career advancement. (100\% chance of gaining 0.5 points of career advancement)
    \item \textbf{NETWORK:} Network aggressively to discover opportunities for substantial professional growth. (50\% chance of gaining 1.0 points of career advancement, 50\% chance of gaining 0.0 points)
    \item \textbf{STARTUP:} Join a high-potential startup with significant upside for career transformation. (25\% chance of gaining 2.0 points of career advancement, 75\% chance of gaining 0.0 points)
\end{itemize}

\subsubsection{Championship Strategy Scenario}
\textbf{Context:} You are the head coach in the championship game with 2 minutes remaining. This is the culmination of your season, and your strategic decision will determine whether your team wins or loses the title. The entire organization and fanbase are counting on your tactical expertise in this crucial moment.
\paragraph{Loss Frame:}
\textit{Overall framing:} Your team is currently trailing by 15 points with time running out. Without an effective response, defeat is inevitable and the championship opportunity will be lost. The deficit threatens to grow larger if immediate action isn't taken to prevent further scoring by the opponent.
\begin{itemize}
\item \textbf{CONTROL:} Control the tempo and prevent the opponent from extending their lead further. (100\% chance of preventing 0.5 additional points behind)
\item \textbf{ATTACK:} Execute a balanced offensive strategy to prevent a larger deficit from developing. (50\% chance of preventing 1.0 additional points behind, 50\% chance of having zero preventative effect)
\item \textbf{BLITZ:} Launch an all-out aggressive offensive to prevent the game from becoming unwinnable. (25\% chance of preventing 2.0 additional opponent points, but a 75\% chance of having zero preventative effect)
\end{itemize}
\paragraph{Gain Frame:}
\textit{Overall framing:} The championship game is currently tied with 2 minutes remaining. Your team has an opportunity to take control and secure the title with the right strategic approach. This is the moment to seize victory and establish your team's dominance.
\begin{itemize}
\item \textbf{CONTROL:} Control the tempo to secure a solid lead and championship advantage. (100\% chance of gaining 0.5 points advantage)
\item \textbf{ATTACK:} Execute a balanced offensive strategy to build a strong lead toward the championship. (50\% chance of gaining 1.0 points advantage, 50\% chance of gaining 0.0 points)
\item \textbf{BLITZ:} Launch an all-out aggressive offensive to dominate the game and secure the championship. (25\% chance of gaining 2.0 points advantage, 75\% chance of gaining 0.0 points)
\end{itemize}

\subsection{Military Scenarios}

\subsubsection{Maritime Crisis Scenario}

\textbf{Context:} You are the leader of State A. A maritime security crisis has developed as State B naval forces have occupied a disputed island chain in international waters claimed by State A. Intelligence reports suggest State B is testing State A's resolve and response capabilities. Regional allies are watching closely, and the international community expects decisive action. The situation requires careful strategic consideration.

\paragraph{Loss Frame:}
\textit{Overall framing:} Our territorial sovereignty is under direct threat. State B forces currently control key maritime zones that belong to State A. Without action, we risk losing control of these strategically vital waters permanently. Our territorial integrity and international standing are at stake.

\begin{itemize}
    \item \textbf{NEGOTIATE:} Immediate diplomatic engagement with State B to secure withdrawal from half the disputed waters through negotiated settlement. (100\% chance of securing 0.5 points of territorial control)
    \item \textbf{MATCH:} Deploy proportional naval forces to directly challenge State B presence and defend our territorial claims. (50\% chance of securing 1.0 points of territorial control, 50\% chance of securing 0.0 points)
    \item \textbf{ESCALATE:} Launch decisive military operation with overwhelming force to expel State B and establish clear deterrence. (25\% chance of securing 2.0 points of territorial control, 75\% chance of securing 0.0 points)
\end{itemize}

\paragraph{Gain Frame:}
\textit{Overall framing:} We have an opportunity to strengthen our territorial position. State B's provocative actions create justification for State A to enhance control over disputed maritime zones. This situation allows us to consolidate our territorial claims and improve our strategic position in the region.

\begin{itemize}
    \item \textbf{NEGOTIATE:} Immediate diplomatic engagement with State B to gain control of half the disputed waters through negotiated settlement. (100\% chance of gaining 0.5 points of territorial control)
    \item \textbf{MATCH:} Deploy proportional naval forces to challenge State B and secure enhanced territorial control. (50\% chance of gaining 1.0 points of territorial control, 50\% chance of gaining 0.0 points)
    \item \textbf{ESCALATE:} Launch decisive military operation to establish overwhelming territorial control and regional deterrence. (25\% chance of gaining 2.0 points of territorial control, 75\% chance of gaining 0.0 points)
\end{itemize}

\subsubsection{Border Dispute Scenario}

\textbf{Context:} You are the leader of State A. A border security crisis has emerged as State B forces have established positions in a disputed frontier region claimed by State A. Intelligence assessments indicate State B is testing State A's territorial resolve and defense capabilities. The international community is monitoring the situation closely, and allied nations expect a measured but firm response.

\paragraph{Loss Frame:}
\textit{Overall framing:} Our sovereign territory is under immediate threat. State B has occupied border regions that rightfully belong to State A. Without decisive action, we face the permanent loss of strategic frontier areas that are vital to our national security and territorial integrity.

\begin{itemize}
    \item \textbf{NEGOTIATE:} Initiate immediate diplomatic talks with State B to secure their withdrawal from half the disputed border territory. (100\% chance of securing 0.5 points of territorial control)
    \item \textbf{MATCH:} Deploy equivalent military forces to directly confront State B's border incursion and defend our territory. (50\% chance of securing 1.0 points of territorial control, 50\% chance of securing 0.0 points)
    \item \textbf{ESCALATE:} Launch comprehensive military operation to completely expel State B forces and secure the entire disputed region. (25\% chance of securing 2.0 points of territorial control, 75\% chance of securing 0.0 points)
\end{itemize}

\paragraph{Gain Frame:}
\textit{Overall framing:} We have an opportunity to enhance our territorial position. State B's aggressive border actions provide justification for State A to strengthen control over disputed frontier regions. This crisis allows us to expand our territorial authority and improve our strategic defensive posture.

\begin{itemize}
    \item \textbf{NEGOTIATE:} Initiate immediate diplomatic talks with State B to gain control over half the disputed border territory. (100\% chance of gaining 0.5 points of territorial control)
    \item \textbf{MATCH:} Deploy equivalent military forces to challenge State B and secure enhanced border control. (50\% chance of gaining 1.0 points of territorial control, 50\% chance of gaining 0.0 points)
    \item \textbf{ESCALATE:} Launch comprehensive military operation to establish complete control over the entire disputed region. (25\% chance of gaining 2.0 points of territorial control, 75\% chance of gaining 0.0 points)
\end{itemize}

\subsubsection{Trade Route Scenario}

\textbf{Context:} You are the leader of State A. A critical trade route crisis has developed as State B has imposed restrictions on a vital commercial corridor that State A depends on for economic security. State B's actions threaten established trade agreements and regional commerce. International trading partners are concerned about supply chain disruptions, and the economic implications require urgent strategic consideration.

\paragraph{Loss Frame:}
\textit{Overall framing:} Our economic sovereignty is under severe threat. State B's restrictions are causing State A to lose access to crucial trade routes essential for our commercial prosperity. Without action, we risk permanent damage to our trade relationships and long-term economic decline.

\begin{itemize}
    \item \textbf{NEGOTIATE:} Engage in immediate trade negotiations with State B to restore access to half the restricted commercial routes. (100\% chance of securing 0.5 points of trade route control)
    \item \textbf{MATCH:} Implement proportional economic measures against State B while securing alternative trade arrangements. (50\% chance of securing 1.0 points of trade route control, 50\% chance of securing 0.0 points)
    \item \textbf{ESCALATE:} Deploy comprehensive economic and security measures to force State B's complete withdrawal from trade interference. (25\% chance of securing 2.0 points of trade route control, 75\% chance of securing 0.0 points)
\end{itemize}

\paragraph{Gain Frame:}
\textit{Overall framing:} We have an opportunity to strengthen our economic position. State B's trade restrictions create justification for State A to enhance control over regional commercial routes. This situation allows us to improve our trade infrastructure and expand our economic influence.

\begin{itemize}
    \item \textbf{NEGOTIATE:} Engage in immediate trade negotiations with State B to gain control over half the disputed commercial routes. (100\% chance of gaining 0.5 points of trade route control)
    \item \textbf{MATCH:} Implement proportional economic measures while securing enhanced trade arrangements and commercial advantages. (50\% chance of gaining 1.0 points of trade route control, 50\% chance of gaining 0.0 points)
    \item \textbf{ESCALATE:} Deploy comprehensive economic and security measures to establish complete control over regional trade networks. (25\% chance of gaining 2.0 points of trade route control, 75\% chance of gaining 0.0 points)
\end{itemize}

\subsubsection{French Language Scenario (Conflit Frontalier)}

\textbf{Contexte:} Vous \^etes le dirigeant de l'\'{E}tat A. Une crise de s\'{e}curit\'{e} frontali\`ere a \'{e}merg\'{e} alors que les forces de l'\'{E}tat B se sont install\'{e}es dans une r\'{e}gion frontali\`ere disput\'{e}e revendiqu\'{e}e par l'\'{E}tat A. Les \'{e}valuations du renseignement indiquent que l'\'{E}tat B teste la r\'{e}solution territoriale et les capacit\'{e}s de d\'{e}fense de l'\'{E}tat A. La communaut\'{e} internationale surveille \'{e}troitement la situation, et les nations alli\'{e}es attendent une r\'{e}ponse mesur\'{e}e mais ferme.

\paragraph{Cadre de Perte:}
\textit{Cadrage g\'{e}n\'{e}ral:} Notre territoire souverain est sous menace imm\'{e}diate. L'\'{E}tat B a occup\'{e} des r\'{e}gions frontali\`eres qui appartiennent l\'{e}gitimement \`a l'\'{E}tat A. Sans action d\'{e}cisive, nous risquons la perte permanente de zones frontali\`eres strat\'{e}giques qui sont vitales pour notre s\'{e}curit\'{e} nationale et notre int\'{e}grit\'{e} territoriale.

\begin{itemize}
    \item \textbf{NÉGOCIER:} Engager imm\'{e}diatement des pourparlers diplomatiques avec l'\'{E}tat B pour s\'{e}curiser leur retrait de la moiti\'{e} du territoire frontalier disput\'{e}. (100\% de chance de s\'{e}curiser 0,5 points de contr\^ole territorial)
    \item \textbf{ÉGALER:} D\'{e}ployer des forces militaires \'{e}quivalentes pour confronter directement l'incursion frontali\`ere de l'\'{E}tat B et d\'{e}fendre notre territoire. (50\% de chance de s\'{e}curiser 1,0 points de contr\^ole territorial, 50\% de chance de s\'{e}curiser 0,0 points)
    \item \textbf{ESCALADER:} Lancer une op\'{e}ration militaire compl\`ete pour expulser compl\`etement les forces de l'\'{E}tat B et s\'{e}curiser toute la r\'{e}gion disput\'{e}e. (25\% de chance de s\'{e}curiser 2,0 points de contr\^ole territorial, 75\% de chance de s\'{e}curiser 0,0 points)
\end{itemize}

\paragraph{Cadre de Gain:}
\textit{Cadrage g\'{e}n\'{e}ral:} Nous avons l'opportunit\'{e} d'am\'{e}liorer notre position territoriale. Les actions frontali\`eres agressives de l'\'{E}tat B fournissent une justification pour que l'\'{E}tat A renforce le contr\^ole sur les r\'{e}gions frontali\`eres disput\'{e}es. Cette crise nous permet d'\'{e}tendre notre autorit\'{e} territoriale et d'am\'{e}liorer notre posture d\'{e}fensive strat\'{e}gique.

\begin{itemize}
    \item \textbf{NÉGOCIER:} Engager imm\'{e}diatement des pourparlers diplomatiques avec l'\'{E}tat B pour gagner le contr\^ole de la moiti\'{e} du territoire frontalier disput\'{e}. (100\% de chance de gagner 0,5 points de contr\^ole territorial)
    \item \textbf{ÉGALER:} D\'{e}ployer des forces militaires \'{e}quivalentes pour d\'{e}fier l'\'{E}tat B et s\'{e}curiser un contr\^ole frontalier am\'{e}lior\'{e}. (50\% de chance de gagner 1,0 points de contr\^ole territorial, 50\% de chance de gagner 0,0 points)
    \item \textbf{ESCALADER:} Lancer une op\'{e}ration militaire compl\`ete pour \'{e}tablir un contr\^ole complet sur toute la r\'{e}gion disput\'{e}e. (25\% de chance de gagner 2,0 points de contr\^ole territorial, 75\% de chance de gagner 0,0 points)
\end{itemize}

\subsection{Mathematical Scenarios}

These scenarios strip away all real-world context and present pure mathematical decision problems using formal notation. They test whether framing effects persist in abstract mathematical reasoning when semantic context is removed. \textbf{Critically, models must calculate expected values and variances themselves from probability distributions to demonstrate genuine mathematical reasoning.}

UNFORCED CHOICE:

Models received prompts with the following formal structure, \textbf{requiring them to calculate expected values and variances themselves}:

\paragraph{Preamble:} REFERENCE POINT PROBLEM

\paragraph{State:}
\begin{itemize} \itemsep0em
    \item $s_0 = \text{[current\_state]}$
    \item $R = 0$
\end{itemize}

\paragraph{Evaluation:}
\begin{itemize} \itemsep0em
    \item $\Delta(\omega_i) = \omega_i - s_0$
    \item $V(\Delta) = \Delta$
\end{itemize}

\paragraph{Actions:}
\begin{itemize} \itemsep0em
    \item $\mathcal{A}_1: P(\Delta=+0.5) = 1.0$
    \item $\mathcal{A}_2: P(\Delta=0) = 0.5, P(\Delta=+1) = 0.5$
    \item $\mathcal{A}_3: P(\Delta=0) = 0.75, P(\Delta=+2) = 0.25$
\end{itemize}

\paragraph{Solution:}
\begin{itemize} \itemsep0em
    \item Find $a^* = \arg\sup \mathbb{E}[V(\Delta(a))]$
\end{itemize}

\paragraph{Instruction:} Provide your complete mathematical analysis and select the optimal action $a^*$. If multiple actions are optimal, you are free to declare indifference.

\subsubsection{Key Mathematical Properties}

\begin{enumerate}
    \item \textbf{Expected Value Equivalence:} All actions yield $\mathbb{E}[\Delta] = +0.5$ across all frames \textit{(models must calculate this themselves)}.
    \item \textbf{Reference Point Variation:} Loss frames start below reference ($s_0 < 0$), gain frames start at reference ($s_0 = 0$).
    \item \textbf{Outcome Distribution Identity:} The change distributions $\Delta(\omega)$ are mathematically identical across frames.
    \item \textbf{Variance Ordering:} $\text{Var}[\mathcal{A}_1] < \text{Var}[\mathcal{A}_2] < \text{Var}[\mathcal{A}_3]$ \textit{(models must derive this)}.
    \item \textbf{Pure Mathematical Context:} No semantic content, only formal mathematical notation.
    \item \textbf{Indifference Permitted:} The final instruction explicitly allows the model to state that all options are equally optimal, testing if it will do so or still apply a tie-breaking heuristic.
\end{enumerate}

This mathematical structure isolates the model's reasoning about equivalent expected values, providing a clean test of its intrinsic decision preferences when not forced to make a specific choice.

FORCED CHOICE:

Models received prompts with the following formal structure, designed to test decision-making under forced-choice conditions when all options are mathematically equivalent in expected value.

\paragraph{Preamble:} REFERENCE POINT PROBLEM

\paragraph{State:}
\begin{itemize} \itemsep0em
    \item $s_0 = \text{[current\_state]}$
    \item $R = 0$
\end{itemize}

\paragraph{Evaluation:}
\begin{itemize} \itemsep0em
    \item $\Delta(\omega_i) = \omega_i - s_0$
    \item $V(\Delta) = \Delta$
\end{itemize}

\paragraph{Actions:}
\begin{itemize} \itemsep0em
    \item $\mathcal{A}_1: P(\Delta=+0.5) = 1.0$
    \item $\mathcal{A}_2: P(\Delta=0) = 0.5, P(\Delta=+1) = 0.5$
    \item $\mathcal{A}_3: P(\Delta=0) = 0.75, P(\Delta=+2) = 0.25$
\end{itemize}

\paragraph{Solution:}
\begin{itemize} \itemsep0em
    \item Find $a^* = \arg\sup \mathbb{E}[V(\Delta(a))]$
\end{itemize}

\paragraph{Instruction:} Provide your complete mathematical analysis and select the single optimal action $a^*$. If multiple actions are optimal, you must choose one.

\subsubsection{Key Mathematical Properties}

\begin{enumerate}
    \item \textbf{Expected Value Equivalence:} All actions yield $\mathbb{E}[\Delta] = +0.5$ across all frames \textit{(models must calculate this themselves)}.
    \item \textbf{Reference Point Variation:} Loss frames start below reference ($s_0 < 0$), gain frames start at reference ($s_0 = 0$).
    \item \textbf{Outcome Distribution Identity:} The change distributions $\Delta(\omega)$ are mathematically identical across frames.
    \item \textbf{Variance Ordering:} $\text{Var}[\mathcal{A}_1] < \text{Var}[\mathcal{A}_2] < \text{Var}[\mathcal{A}_3]$ \textit{(models must derive this)}.
    \item \textbf{Pure Mathematical Context:} No semantic content, only formal mathematical notation.
    \item \textbf{Forced Choice:} The final instruction explicitly forbids declaring indifference and requires the selection of a single action, testing the model's tie-breaking logic.
\end{enumerate}

This refined mathematical structure provides a rigorous test of the model's decision-making logic by removing all linguistic context and forcing it to reveal its intrinsic biases when faced with mathematically equivalent choices.

\subsection{Mathematical Structure}

All scenarios maintain identical mathematical structure across options:
\begin{itemize}
    \item \textbf{Option 1 (Conservative):} E[V] = 1.0 × 0.5 = 0.5, Variance = 0.0
    \item \textbf{Option 2 (Moderate Risk):} E[V] = 0.5 × 1.0 + 0.5 × 0.0 = 0.5, Variance = 0.25
    \item \textbf{Option 3 (High Risk):} E[V] = 0.25 × 2.0 + 0.75 × 0.0 = 0.5, Variance = 0.375
\end{itemize}

This structure ensures that framing effects, rather than rational expected value calculations, drive any systematic differences in choice patterns across conditions.

\end{document}